\documentclass[letterpaper]{article} 
\usepackage{aaai25}  
\usepackage{times}  
\usepackage{helvet}  
\usepackage{courier}  
\usepackage[hyphens]{url}  
\usepackage{graphicx} 
\usepackage{natbib}  
\usepackage{caption} 
\usepackage{algorithm}
\usepackage{algorithmic}

\usepackage{newfloat}
\usepackage{listings}

\DeclareCaptionStyle{ruled}{labelfont=normalfont,labelsep=colon,strut=off} 
\floatstyle{ruled}
\newfloat{listing}{tb}{lst}{}
\floatname{listing}{Listing}
\usepackage{subcaption}
\usepackage{amsmath}
\usepackage{amssymb}
\usepackage{multirow}
\usepackage{array}
\usepackage{enumitem}
\usepackage{booktabs}
\usepackage{appendix}
\usepackage{makecell}

\title{StructSR: Refuse Spurious Details in Real-World Image Super-Resolution}
\author{
    Yachao Li\textsuperscript{\rm 1},
    Dong Liang\textsuperscript{\rm 1}\equalcontrib,
    Tianyu Ding\textsuperscript{\rm 2},
    Sheng-Jun Huang\textsuperscript{\rm 1}
}
\affiliations{
    \textsuperscript{\rm 1}College of Computer Science and Technology, Nanjing University of Aeronautics and Astronautics, MIIT Key Laboratory of Pattern Analysis and Machine Intelligence, Collaborative Innovation Center of Novel Software Technology and Industrialization, Nanjing 211106, China.\\
    \textsuperscript{\rm 2}Microsoft, Washington, USA\\

    \{liyachao, liangdong, huangsj\}@nuaa.edu.cn, tianyuding@microsoft.com
}

\usepackage{bibentry}

\begin{document}

\maketitle

\begin{abstract}
Diffusion-based models have shown great promise in real-world image super-resolution (Real-ISR), but often generate content with structural errors and spurious texture details due to the empirical priors and illusions of these models. To address this issue, we introduce StructSR, a simple, effective, and plug-and-play method that enhances structural fidelity and suppresses spurious details for diffusion-based Real-ISR. StructSR operates without the need for additional fine-tuning, external model priors, or high-level semantic knowledge. At its core is the Structure-Aware Screening (SAS) mechanism, which identifies the image with the highest structural similarity to the low-resolution (LR) input in the early inference stage, allowing us to leverage it as a historical structure knowledge to suppress the generation of spurious details. By intervening in the diffusion inference process, StructSR seamlessly integrates with existing diffusion-based Real-ISR models. Our experimental results demonstrate that StructSR significantly improves the fidelity of structure and texture, improving the PSNR and SSIM metrics by an average of 5.27\% and 9.36\% on a synthetic dataset (DIV2K-Val) and 4.13\% and 8.64\% on two real-world datasets (RealSR and DRealSR) when integrated with four state-of-the-art diffusion-based Real-ISR methods. 
\end{abstract}

\begin{links}
\link{Code}{https://github.com/LYCEXE/StructSR}
\end{links}

\section{Introduction}
Image super-resolution aims to reconstruct high-resolution (HR) images from their low-resolution (LR) counterparts. Traditional image super-resolution methods often rely on simplistic assumptions about degradation (e.g., Gaussian noise and bicubic downsampling) and design methods tailored to these degradation models~\cite{chen2021pre,chen2023activating,dong2014learning,liang2021swinir}. However, their ability to generalize to real-world images with complex degradation is limited. To address the challenges of real-world image super-resolution (Real-ISR), methods~\cite{zhang2021designing,wang2021real} have employed Generative Adversarial Networks (GANs)~\cite{goodfellow2020generative} to generate HR images from LR images collected from the real world. However, these methods rely heavily on specific paired training data and generally lack the capability to generate realistic texture details for Real-ISR.

\begin{figure}[t]
\centering
\includegraphics[width=\columnwidth]{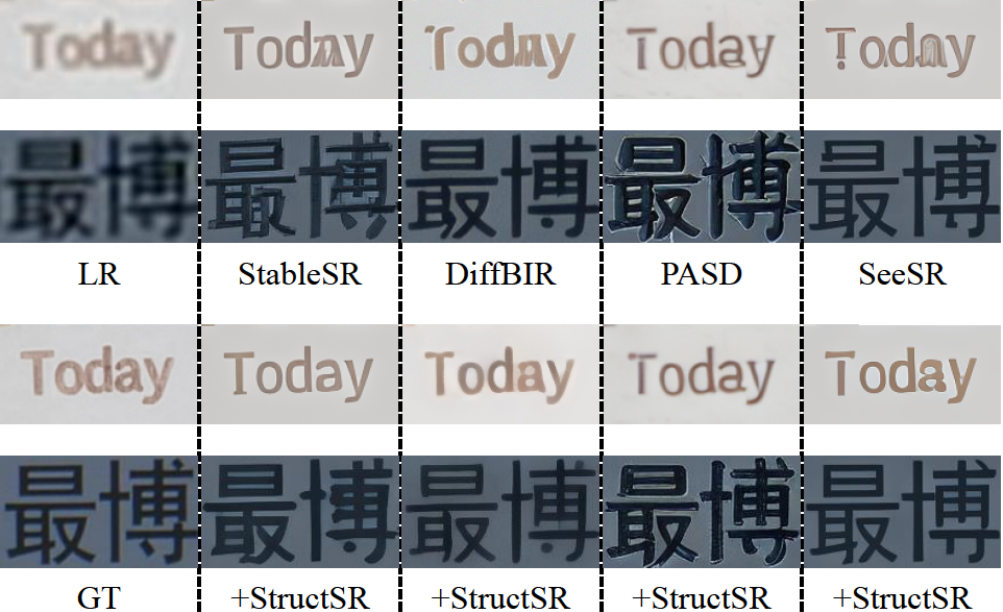}
\caption{Comparison of diffusion-based Real-ISR methods with and without StructSR integration. The original methods generate spurious details in both English letters and Chinese characters. Integration with StructSR significantly reduces these artifacts, resulting in more accurate reconstruction.
}
\label{fig1}
\end{figure}

Current diffusion-based Real-ISR methods typically utilize pre-trained text-to-image models, e.g., Stable Diffusion~\cite{rombach2022high} as priors and fine-tune them using real-world image super-resolution datasets. These methods~\cite{wang2023exploiting, lin2023diffbir,yang2023pixel,wu2024seesr} have shown remarkable proficiency in generating realistic image details. However, they often struggle with maintaining structural fidelity, and the generated details may be spurious with the real semantics due to the empirical priors and illusions of these models, as illustrated in Fig. \ref{fig1}.

We study the changes in the image structure during the inference process of the diffusion-based Real-ISR method by using structural similarity (SSIM)~\cite{wang2004image} between the temporal reconstructed image and the input LR image (Details see Methodology Section). Our analysis reveals that, the temporal reconstructed images with high SSIM values can be used to guide the inference process for eliminating structural errors and suppressing spurious details. Based on this finding, we propose a structure-aware Real-ISR ({StructSR}) method, a simple, effective, and plug-and-play method that enhances structural fidelity for diffusion-based Real-ISR. StructSR consists of three modules: Structure-Aware Screening (SAS), Structure Condition Embedding (SCE), and Image Details Embedding (IDE). SAS identifies the image with the highest structural similarity to the LR image in the early inference stage. SCE uses structural embedding from SAS to guide the inference in conjunction with the LR image, promoting the generation of high-fidelity structural information. IDE inserts the structural embedding into the clean latent image at each timestep, according to the degradation degree of the LR image, to suppress possible spurious details. Extensive experiments demonstrate the effectiveness of StructSR in enhancing the structural fidelity of diffusion-based Real-ISR methods while effectively suppressing potential spurious details.

In summary, we make the following contributions:
\begin{itemize}[leftmargin=*]
\item [$\bullet$] We propose StructSR, to fully leverage the temporal reconstructed images during the inference process to enhance the structural fidelity of the diffusion-based Real-ISR methods, without introducing any excess fine-tuning, external models’ prior, or high-level semantic knowledge.

\item [$\bullet$] We introduce SAS, SCE, and IDE to interactively update the predicted noise and clean images according to the degradation degree of the LR image during inference, enabling a plug-and-play intervention generation process for diffusion-based Real-ISR methods while suppressing potential spurious structure and texture details.

\item [$\bullet$]  We demonstrate through experiments that the proposed StructSR consistently improves the structural fidelity of diffusion-based and GAN-based Real-ISR methods.

\end{itemize}

\section{Related Work}
\textbf{Real-ISR.} Starting with SRCNN~\cite{dong2014learning}, learning-based ISR has gained significant popularity. Numerous methods~\cite{chen2021pre,chen2023activating,dat,san, edsr} focusing on deep model design have been proposed. Some methods have explored more complex degradation models to approximate real-world degradation. BSRGAN~\cite{zhang2021designing} introduces a randomly shuffled degradation modeling, while Real-ESRGAN~\cite{wang2021real} adopts a high-order degradation modeling. Both BSRGAN and Real-ESRGAN employ the Generative Adversarial Networks (GANs)~\cite{goodfellow2020generative} to reconstruct desired HR images using training samples with more realistic degradations. Although these methods generate high-fidelity structures, the training of GANs often suffers from instability, leading to unnatural artifacts in the Real-ISR outputs. Subsequent approaches like FeMaSR~\cite{chen2022femasr} and LDL~\cite{liang2022details} have been developed to mitigate this issue of artifacts. 
However, they still rely heavily on specific paired training data and lack the capability to generate realistic details.

\noindent \textbf{Diffusion Probabilistic Models.} The diffusion model is based on non-equilibrium thermodynamic theory~\cite{jarzynski1997equilibrium} and the Monte Carlo principle~\cite{neal2001annealed}. It utilizes a diffusion process to iteratively sample from the data distribution, allowing it to capture underlying structures and features in high-dimensional spaces.~\citeauthor{dhariwal2021diffusion} demonstrated that the diffusion model has generation capabilities beyond GAN. Research on accelerated samplers has significantly improved the efficiency of diffusion models, such as DDIM~\cite{rombach2022high}. Large-scale latent space-based pre-trained text-to-image (T2I) diffusion models further improved the performance of the diffusion model by moving it from pixel space to latent space, such as Imagen~\cite{imagen}. Meanwhile, the T2I diffusion model is gradually used in image restoration~\cite{meng2021sdedit,zhang2023adding}, video generation~\cite{singer2022make,wu2023tune}, 3D content generation~\cite{lin2023magic3d,wang2023prolificdreamer}, etc.

\noindent \textbf{Diffusion-based Real-ISR.} Most early attempts~\cite{kawar2022denoising,sr3,wang2022zero,fei2023generative} to utilize diffusion models for ISR were based on the assumption of simplistic degradations. 
Recently, researchers have turned to pre-trained text-to-image (T2I) models like Stable Diffusion~\cite{rombach2022high} (SD) to tackle the challenges of Real-ISR, which is trained on large-scale image-text pairs datasets. StableSR~\cite{wang2023exploiting} refines SD through fine-tuning with a time-aware encoder. DiffBIR~\cite{lin2023diffbir} employs a two-stage strategy, initially preprocessing the image as an initial estimate and subsequently fine-tuning SD to enhance image details. PASD~\cite{yang2023pixel} extracts text prompts from LR images and combines them with a pre-trained SD model using a pixel-aware cross-attention module. To further enhance the semantic perception of Real-ISR, SeeSR~\cite{wu2024seesr} introduces degradation-aware sematic prompts, combining with soft labels to jointly guide the diffusion process.

The above diffusion-based Real-ISR methods ignore the negative impact of the model's empirical prior, resulting in structural errors and spurious details. Since diffusion models are often criticized for their training and inference efficiency, instead of introducing additional degradation-aware sematic prompts (like SeeSR), an additional pre-processing stage (like DiffBIR), or better training materials for fine-tuning, we plan to explore the characteristics of the diffusion model during the inference process to solve this problem.

\section{Methodology}
\subsection{Basic Definition in Diffusion-based Real-ISR}
The diffusion-based inference defines the total inference timesteps $T$ and randomly samples a noisy latent image $Z_T$ from a normal distribution $Z_T \sim \mathcal{N}(0,\textbf{I})$ as the initialization. Given the LR image $I_{LR}$, it is mapped to the latent space through an encoder $\mathcal{E}$ and uses $\mathcal{E}(I_{LR})$ as the control condition. According to $\mathcal{E}(I_{LR})$, the pre-trained denoiser $\epsilon_\theta$ predicts the noise $\epsilon_t$ at each timestep $t \in [0, T]$ to denoise the noisy latent image $Z_{t}$ and generates the clean latent image $Z_{0|t}$. The next noisy latent image $Z_{t-1}$ is obtained by adding noise to $Z_{0|t}$ according to the specific sampler. After performing the above process $T$ times, the clean latent image $Z_{0}$ is generated and the reconstructed image $I_{HR}$ is the output of decoder $\mathcal{D}$, denoted as $\mathcal{D}(Z_{0})$. A plug-and-play framework can directly guide $\epsilon_t$ and $Z_{0|t}$ to enhance the diffusion-based Real-ISR methods.

\subsection{Role of Structural Similarity in Real-ISR}
We then investigate the changes in structural fidelity of the reconstructed SR image during inference. We utilize the structural similarity (SSIM)~\cite{wang2004image} to measure the structural fidelity of the reconstructed images:
\begin{equation}
   \mathbf{SSIM}(x,y) = \frac{{(2\mu_x\mu_y + o)(2\sigma_{xy} + o)}}{{(\mu_x^2 + \mu_y^2 + o)(\sigma_x^2 + \sigma_y^2 + o)}}
\label{eq:1}
\end{equation}
where $\mu_x$ and $\mu_y$ are the means, $\sigma_x$ and $\sigma_y$ are the standard deviations respectively, $\sigma_{xy}$ is the covariance, and $o$ is constant used to avoid denominators being zero. Specifically, we first prepare LR images with varying degradation degrees by applying a combination of downsampling, Gaussian Kernel blur, and JPEG compression to real-world images. We use StableSR~\cite{wang2023exploiting} for Real-ISR and set the total inference timesteps $T = 200$. At each timestep $t$, the decoder $\mathcal{D}$ generates the reconstructed image according to the clean latent image $Z_{0|t}$, denoted as $\mathcal{D}(Z_{0|t})$. We then resize the LR image $I_{LR}$ by bicubic interpolation to maintain the same size with $\mathcal{D}(Z_{0|t})$, denoted as $SR(I_{LR})$, and calculate the SSIM value between $\mathcal{D}(Z_{0|t})$ and $SR(I_{LR})$ by Eq.~\ref{eq:1}:
\begin{equation}
    S_{t} = \mathbf{SSIM}(\mathcal{D}(Z_{0|t}),SR(I_{LR}))
\label{eq:2}
\end{equation}
where $S_{t}$ represents the calculated SSIM value at timestep $t$.

\begin{figure}[!t]
\centering
\begin{subfigure}{\columnwidth}
        \centering
        \includegraphics[width=\columnwidth]{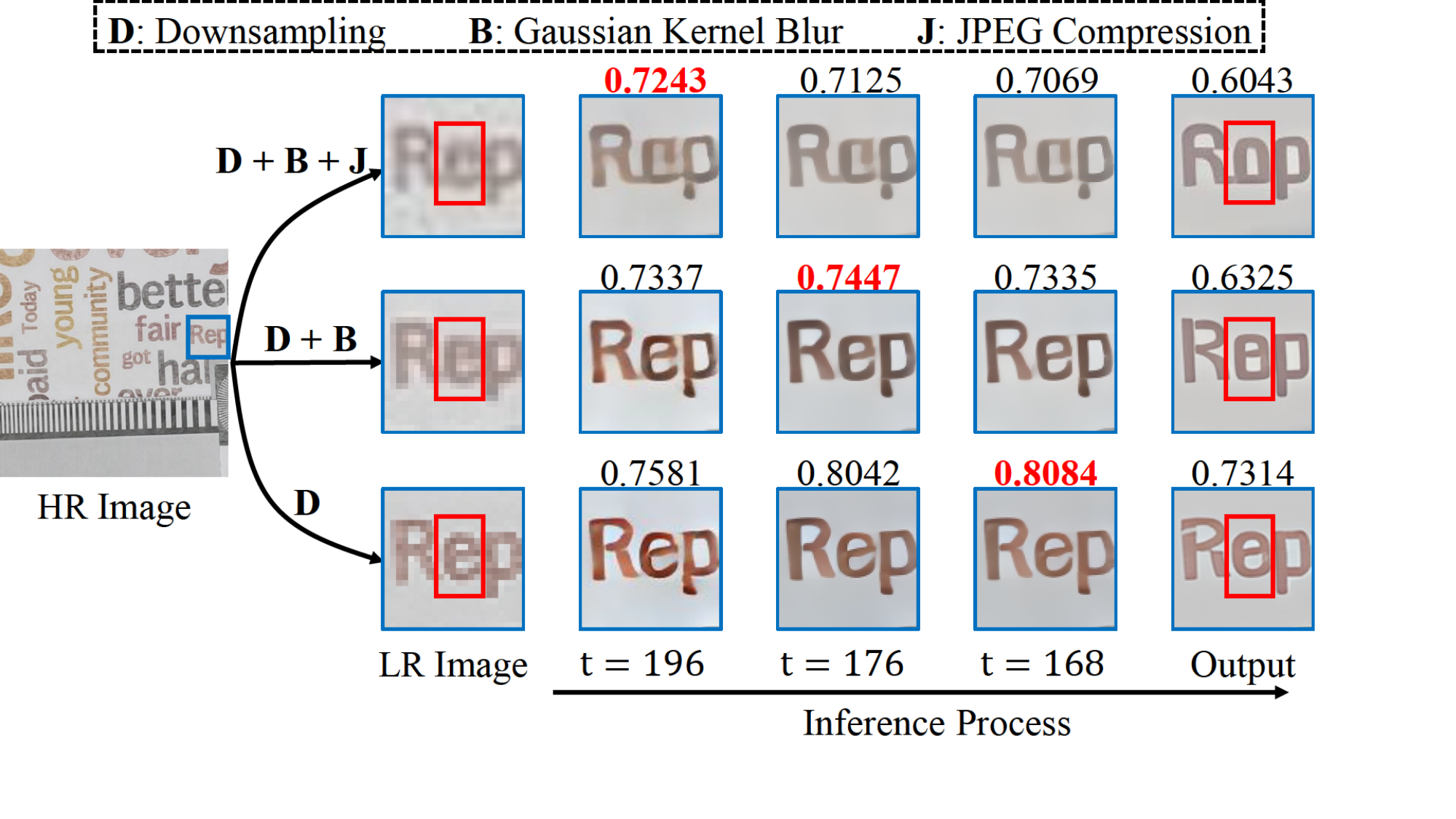}
\end{subfigure}
\hfill
\begin{subfigure}{\columnwidth}
        \centering
        \includegraphics[width=\columnwidth]{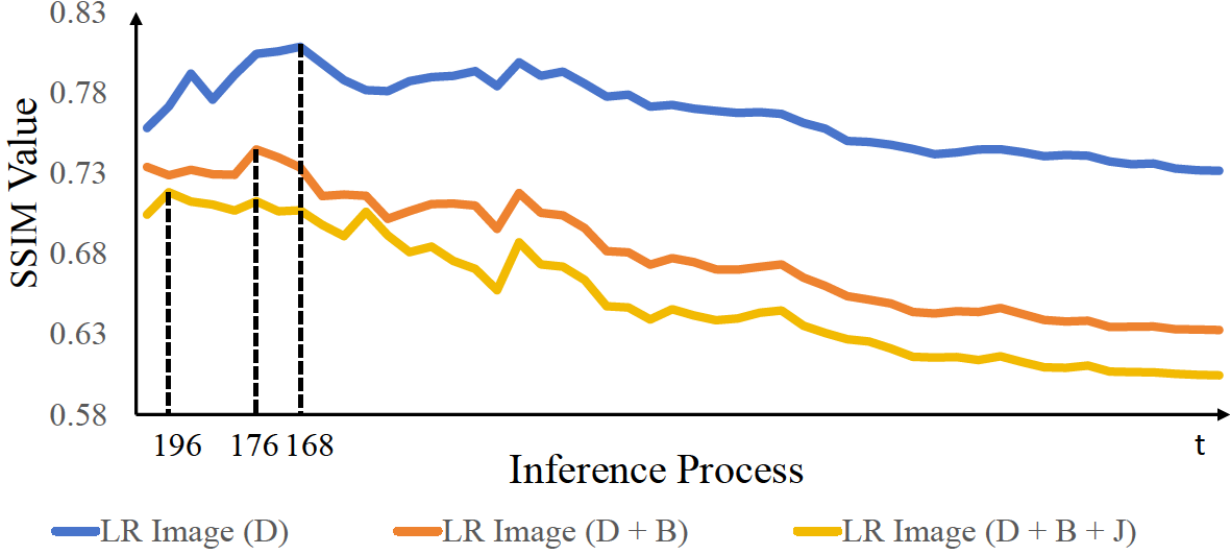}
\end{subfigure}
\caption{Comparison of the structural similarity (SSIM) between LR images with different degradation degrees and their temporal reconstructed images during the inference process. The calculated SSIM values are shown on the top of the reconstructed images, with the maximum SSIM value in red. The red boxes show the issues of structural errors and spurious details. It cannot maintain a stable SSIM, indicating the generation of spurious structure and texture details in the later stage of the inference. }
\label{fig2}
\end{figure}

As shown in Fig.~\ref{fig2}, by comparing the SSIM values and the reconstructed images throughout the inference process, we find that the reconstructed images show the most consistent and clearer structure compared with the LR images in the early inference stage.  
However, the model cannot maintain this clear structure, resulting in decreasing SSIM, indicating the generation of spurious structure and texture details.  
Another recent work DiffBIR~\cite{lin2023diffbir}, uses an additional pre-processing stage to provide clear guidance images by introducing an additional model. 
In our work, we consider screening out the one with the most consistent structure with the LR image to guide the inference process. 
We define the initial $T_{SAS}$ inference timesteps as the early inference stage, and define a structural embedding $Z_{SE}$ to intervene in the clean latent image $Z_{0|t}$. 
Since $S_t$ calculated in the early inference stage reflects the degradation degree of the LR image, it can be used in the structural embedding $Z_{SE}$ to control the guidance strength of the reconstructed image after the inference timesteps $T_{SAS}$.

\begin{figure*}[t]
 \centering
    \begin{subfigure}{\textwidth}
        \centering
        \label{fig3a}
        \includegraphics[width=.9\textwidth]{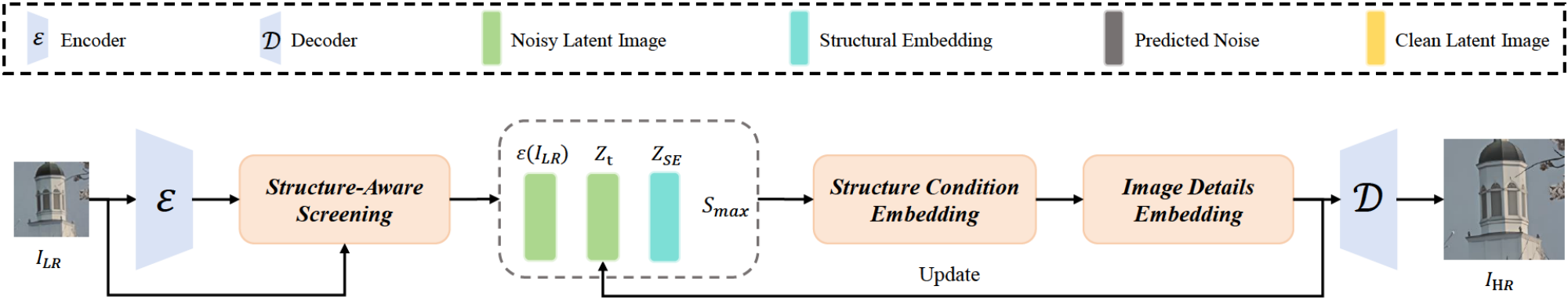}
        \caption{Overview of the proposed StructSR}
    \end{subfigure}
    \begin{subfigure}{0.456\textwidth}
        \centering
        \label{fig3b}
        \includegraphics[width=\textwidth]{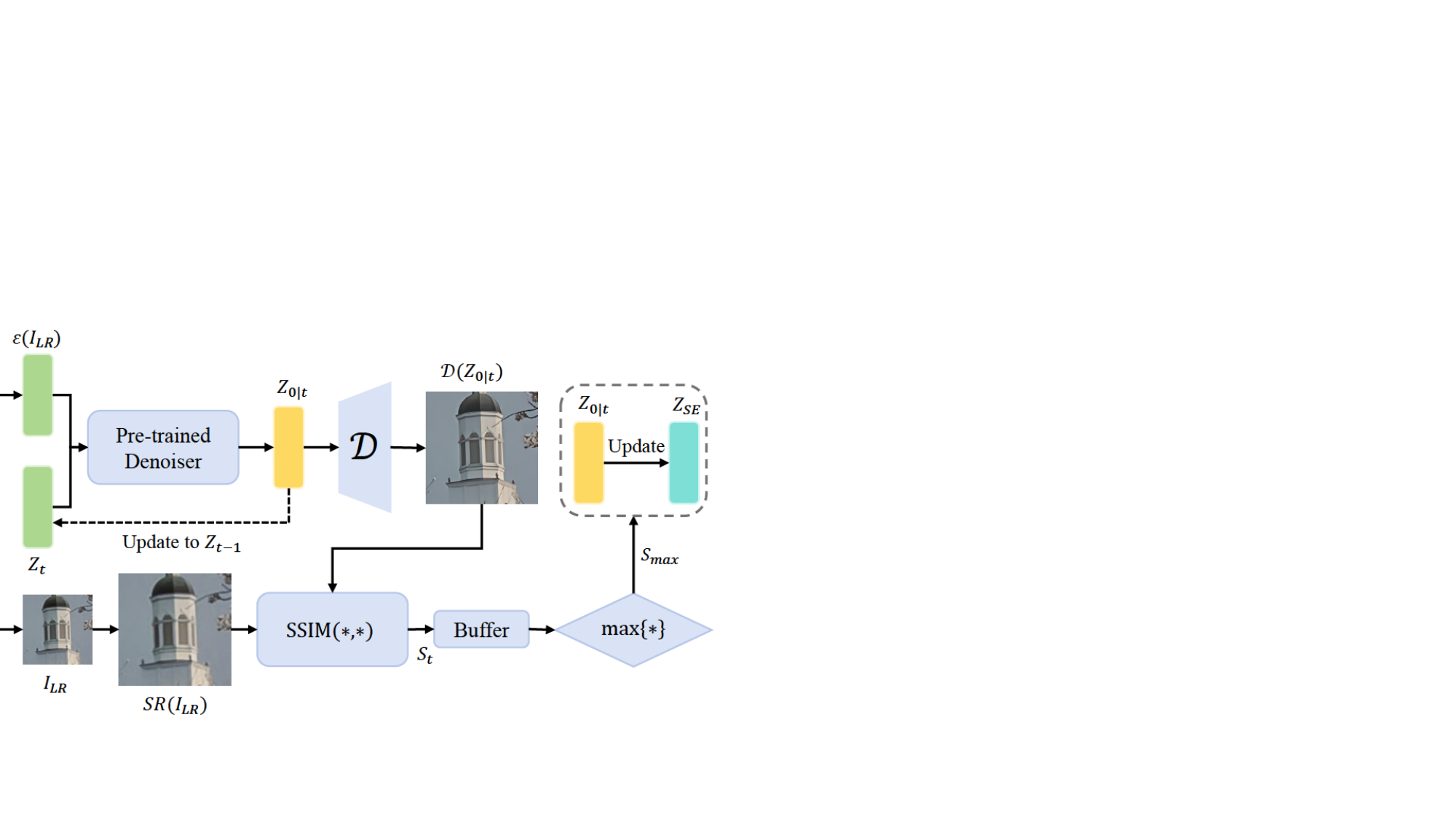}
        \caption{Structure-Aware Screening}
    \end{subfigure}
    \hfill
    \begin{subfigure}{0.324\textwidth}
        \centering
        \label{fig3c}
        \includegraphics[width=\textwidth]{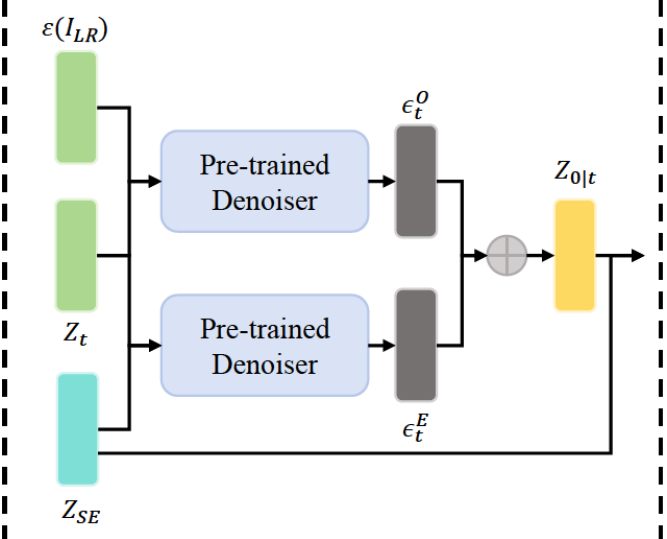}
        \caption{Structure Condition Embedding}
    \end{subfigure}
    \hfill
    \begin{subfigure}{0.21\textwidth}
        \centering
        \label{fig3d}
        \includegraphics[width=\textwidth]{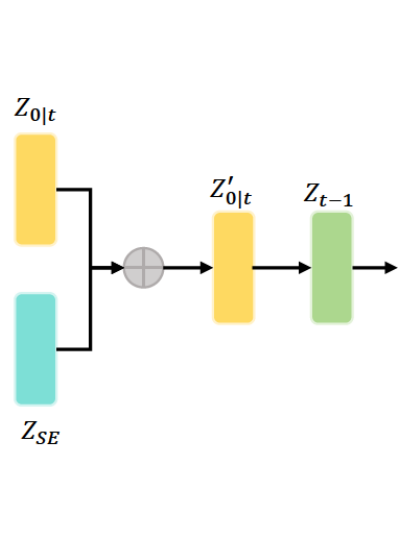}
        \caption{Image Details Embedding}
    \end{subfigure}
    \caption{In the proposed StructSR, the Structure-Aware Screening (SAS)  works in the early inference stage and screens out the structural embedding $Z_{SE}$ with the most consistent and clearer structure compared to the LR image. 
    In the later inference stage, The Structure Condition Embedding (SCE) uses $Z_{SE}$ to guide $\epsilon_t$ in conjunction with the LR image. The Image Details Embedding (IDE) inserts $Z_{SE}$ into the clean latent image $Z_{0|t}$ at each timestep according to the degradation degree.}

\label{fig3}
\end{figure*}

Based on the above observations and considerations, our proposed framework is shown in Fig.~\ref{fig3}(a). 
The structure-aware screening (SAS) works in the early inference stage and screens out the structural embedding $Z_{SE}$ according to $S_t$. The structure condition embedding (SCE) uses the structural embedding $Z_{SE}$ to guide the prediction of $\epsilon_t$. The image details embedding (IDE) inserts the structural embedding $Z_{SE}$ into the clean latent image at each timestep to guide $Z_{0|t}$.

\subsection{Structure-Aware Screening}
Structure-Aware Screening (SAS) is implemented by adding operations based on SSIM in the original inference. As shown in Fig.~\ref{fig3}(b), at each timestep $t$ in the initial $T_{SAS}$ inference timesteps, we use the decoder $\mathcal{D}$ to obtain the reconstructed image $\mathcal{D}(Z_{0|t})$ before updating $Z_{t}$ to $Z_{t-1}$. Then, we use Eq.~\ref{eq:2} to calculate the SSIM value $S_t$ between $\mathcal{D}(Z_{0|t})$ and the LR image $I_{LR}$. We screen out the reconstructed image with the structure that is most consistent with the LR image to ensure the best guidance. Specifically, we use a buffer to store the calculated SSIM values and identify the maximum one from the buffer by $S_{max}=\max\{S_t\}$. If $S_{t} = S_{max}$, we assign this clean latent image $Z_{0|t}$ as the structural embedding $Z_{SE}$,

\begin{equation}
 Z_{SE} = \begin{cases}
Z_{0|t}, & S_{t\in [T-T_{SAS}, T]} = S_{max} \\
Z_{SE}, & \text{otherwise}
\end{cases}
\end{equation}

SAS provides the structural embedding $Z_{SE}$ and the maximum SSIM value $S_{max}$ for intervention in the subsequent inference process.
$S_{max}$ occurs in the early inference stage $t \in [T-T_{SAS}, T]$, which is typically proportional to the original image quality. Better original image quality can ensure that $S_{max}$ occurs later, as shown in Fig.~\ref{fig2}.

\subsection{Structure Condition Embedding }
Structural Conditional Embedding (SCE), as shown in Fig.~\ref{fig3}(c), uses $Z_{SE}$ as the condition of noise prediction to guide $\epsilon_t$ with clearer structural information for addressing the issue of structural errors.
Specifically, in the original inference process, the pre-trained denoiser $\epsilon_\theta$ uses $\mathcal{E}(I_{LR})$ as the control condition for the noise prediction. We defined this predicted noise as $\epsilon^{O}_{t}$. SCE uses $Z_{SE}$ as the additional control condition for $\epsilon_\theta$ and predicts the extra noise, denoted as $\epsilon^{E}_{t}$. SCE uses $\epsilon^{O}_{t}$ to suppress the impact of illusions by constraining $\epsilon^{E}_{t}$ and defines it as follows:

\begin{equation}
 \hat{\epsilon}_{t} = S_{max} \epsilon^{E}_{t} + (1-S_{max}) \epsilon^{O}_{t} 
\end{equation}
where $S_{max}$ is the maximum SSIM value provided by SAS.

The constraint strength provided by $\epsilon^{O}_{t}$ depends on the structural clarity of the reconstructed image corresponding to the control condition $Z_{SE}$. It is directly decided by the degradation degree of the LR image which is reflected by $S_{max}$. Hence, SCE uses $S_{max}$ to control the constraint strength to provide more accurate structural guidance.

\subsection{Image Details Embedding}
Image Details Embedding (IDE), as shown in Fig. \ref{fig3}(d), works in the remaining $T-T_{SAS}$ inference timesteps. Compared with the subsequently reconstructed image, the image corresponding to $Z_{SE}$ contains insufficient details. But it also has the advantage of fewer spurious details and can be used to suppress subsequent spurious details caused by the model's illusions. IDE inserts $Z_{SE}$ into the clean latent image $Z_{0|t}$ at each timestep $t$, where $t \in [0, T-T_{SAS}]$. The insertion process can be expressed as follows: 

\begin{equation}
 Z^{'}_{0|t} = w_t Z_{SE} + (1 - w_t) Z_{0|t}
\end{equation}
Where $w_t$ is the factor that controls the insertion ratio.

The image reconstructed from a severely degraded LR image is severely missing details. Continuously and intensively inserting it into subsequent reconstructed images will cause the final SR image to be too smooth. To avoid this problem, IDE determines the insertion rate according to the degradation degree of the LR image and gradually weakens it. Specifically, the control factor $w_t$ is weakened according to the current timestep $t$, and the remaining inference timesteps $T-T_{SAS}$ as follows:

\begin{equation}
w_t = \frac{S_{max} t}{T-T_{SAS}}
\end{equation}
Where $S_{max}$ is the maximum SSIM value provided by SAS and reflects the degradation degree of the LR image.

\begin{algorithm}[tb]
\caption{StructSR Inference Process}
\label{alg:algorithm}
\textbf{Input}: $I_{LR}$\makebox[1.91cm][l]{}$\triangleright$ LR image\\
\textbf{Parameter}: $T$\makebox[1.48cm][l]{}$\triangleright$ Total inference timesteps\\
\makebox[1.75cm][l]{}$T_{SAS}$\makebox[0.96cm][l]{}$\triangleright$  Inference timesteps with SAS\\
\makebox[1.75cm][l]{}$\mathcal{E}$\makebox[1.57cm][l]{}$\triangleright$ Encoder\\
\makebox[1.75cm][l]{}$\mathcal{D}$\makebox[1.53cm][l]{}$\triangleright$ Decoder\\
\makebox[1.75cm][l]{}$\epsilon_{\theta}$\makebox[1.5cm][l]{}$\triangleright$ Pre-trained denoiser\\
\textbf{Output}: $I_{HR}$
\begin{algorithmic}[1] 
\STATE $Z_T \sim \mathcal{N}(0,\textbf{I})$
\FOR{$t \in [T,...,0]$}
\STATE $\epsilon^{O}_{t}$ = $\epsilon_{\theta}(Z_t,\mathcal{E}(I_{LR}),t)$
\IF {$t>T-T_{SAS}$} 
\STATE $Z_{0|t}=$ Sampler$(Z_t,\epsilon^{O}_{t},t)$
\STATE $S_{t} = \mathbf{SSIM}(\mathcal{D}(Z_{0|t}),SR(I_{LR}))$
\STATE Add $S_{t}$ to Buffer
\STATE $S_{max} = \max\{S_t\}$
\STATE  $Z_{SE} = \begin{cases}
Z_{0|t}, & S_{t} = S_{max} \\
Z_{SE}, & \text{otherwise}
\end{cases} $
\STATE $Z_{t-1}=$ Sampler$(Z_{0|t},\epsilon^{O}_{t},t)$
\ELSE
\STATE $\epsilon^{E}_{t}$ = $\epsilon_{\theta}(Z_t,Z_{SE},t)$
\STATE $\hat{\epsilon}_{t} = S_{max} \epsilon^{E}_{t} + (1-S_{max}) \epsilon^{O}_{t} $
\STATE $Z_{0|t}=$ Sampler$(Z_t,\hat{\epsilon}_{t},t)$
\STATE $w_t = \frac{S_{max} t}{T-T_{SAS}}$
\STATE $Z^{'}_{0|t} = w_t Z_{SE} + (1 - w_t) Z_{0|t}$
\STATE $Z_{t-1}=$ Sampler($Z^{'}_{0|t},\hat{\epsilon}_{t},t)$
\ENDIF
\ENDFOR
\STATE $I_{HR} = \mathcal{D}(Z_{0})$
\STATE \textbf{return} $I_{HR}$
\end{algorithmic}
\end{algorithm}

StructSR exploits the structural information via $S_{max}$ and $Z_{SE}$ during the inference process and interactively updates the predicted noise $\epsilon_t$  and clean latent images $Z_{0|t}$ according to the degradation degree. This enables a plug-and-play intervention inference process for existing diffusion-based Real-ISR methods to suppress potential spurious structure and texture details, while avoiding oversmoothing the real details. 
Algorithm~\ref{alg:algorithm} summarizes the proposed StructSR.

\section{Experiments}
\subsection{Experimental Settings}
\textbf{Test Datasets.} We use one synthetic dataset and two real-world datasets to comprehensively evaluate StructSR. Following the pipeline in StableSR~\cite{wang2023exploiting}, first, we randomly crop 3K patches (resolution: 512×512) from the DIV2K validation set~\cite{div2k} and degrade them as synthetic images, named  DIV2K-Val. Then, we use two real-world datasets, RealSR~\cite{realsr} and DRealSR~\cite{drealsr}, to center-crop LR images (128×128) as real-world images.

\noindent \textbf{Evaluation Metrics.} In order to provide a comprehensive and holistic assessment of the performance of different methods, we employ a range of reference and no-reference metrics. PSNR and SSIM~\cite{wang2004image} (calculated on the Y channel in YCbCr space) are reference-based structural fidelity measures, while LPIPS~\cite{lpips} is the reference-based perceptual quality metric. MUSIQ~\cite{musiq} and CLIPIQA~\cite{clipiqa} are two no-reference image quality metrics.

\noindent \textbf{Compared Methods and Implementation Details.} We adopt four state-of-the-art diffusion-based Real-ISR methods StableSR~\cite{wang2023exploiting}, DiffBIR~\cite{lin2023diffbir}, PASD~\cite{yang2023pixel}, and SeeSR~\cite{wu2024seesr} as baselines and test them using publicly released codes and models. The model parameters and samplers are set according to the original paper. 

For StructSR, we set $T_{SAS}=0.3\ T$ as the inference timesteps for screening and intervene in the inference process according to the sampler. We present the \textbf{ablation study} on $T_{SAS}$, SCE, and IDE in the supplementary material.

 To provide a comprehensive assessment, we also show the comparison with the GAN-based Real-ISR methods. We choose BSRGAN~\cite{zhang2021designing}, Real-ESRGAN~\cite{wang2021real}, LDL~\cite{liang2022details}, and FeMaSR~\cite{chen2022femasr}. For them, we directly use the publicly released codes and models for testing.

\subsection{Comparison with the State-of-the-Art}
\textbf{Quantitative Comparisons.} We present quantitative comparisons on three synthetic and real datasets in Table \ref{tab_cmp}. We can find the following phenomena. (1) Compared with the original diffusion-based methods without integrating with our StructSR, GAN-based methods have significant advantages in PSNR, SSIM, and LPIPS metrics while showing disadvantages in no-reference metrics MUSIQ and CLIP-IQA. This demonstrates that the GAN-based methods generate images with higher structural fidelity but suffer from the problem of insufficient realistic details. (2) Compared with the four diffusion-based baselines, the methods integrating with our StructSR achieved the best results in PSNR and SSIM on all datasets. Although LPIPS did not achieve optimal results, they are better than the original. (3) After StructSR enhancement, StableSR, PASD, and SeeSR achieved better results on MUSIQ and CLIP-IQA metrics. Although DiffBIR slightly decreased on the MUSIQ and CLIP-IQA, it significantly improved on PSNR and SSIM, especially PSNR, achieving the best results among all methods. In summary, our StructSR significantly enhances the performance of diffusion-based Real-ISR methods across full-reference metrics and marginally improves their performance in no-reference metrics, which demonstrates the effectiveness of StructSR for enhancing structural fidelity and suppressing spurious details.

\begin{table*}[t!]
\setlength{\tabcolsep}{1mm}
\centering
\begin{tabular}{c|c|c|c|c|c|c}
\toprule
Datasets                    & Methods            & PSNR $\uparrow$ & SSIM $\uparrow$ & LPIPS $\downarrow$ & MUSIQ $\uparrow$ & CLIP-IQA $\uparrow$ \\ \hline
\multirow{8}{*}{\makecell{DIV2K-Val \\  (synthetic)}} & BSRGAN           &  23.26    & 0.5907     &  0.3351     &  61.20     &  0.5247       \\  
                            & Real-ESRGAN      &  22.97    & 0.5986     &  0.3112     &  61.07     &  0.5281       \\
                            & FeMaSR           &  21.74    & 0.5536     &  0.3126     &  60.83     &  0.5998       \\
                            & LDL           &  22.17    & 0.5798        &  0.3256     &  60.04     &  0.5179       \\ \cline{2-7} 
                            & StableSR / +StructSR &  21.94 / 23.47 $\uparrow$   & 0.5335 / \textbf{0.6117} $\uparrow$    &  0.3122 / \textbf{0.3004} $\downarrow$    &  65.87 / 66.72 $\uparrow$    &  0.6776 / 0.6816 $\uparrow$       \\
                            & DiffBIR / +StructSR  &  22.31 / \textbf{23.60} $\uparrow$   & 0.5272 / 0.5835 $\uparrow$    &  0.3520 / 0.3286 $\downarrow$    &  65.78 / 65.02 $\downarrow$    &  0.6696 / 0.6558 $\downarrow$       \\
                            & PASD / +StructSR     &  22.31 / 23.35 $\uparrow$   & 0.5656 / 0.6029 $\uparrow$    &  0.3728 / 0.3072 $\downarrow$    &  64.52 / 67.93 $\uparrow$    &  0.6180 / 0.6951 $\uparrow$       \\
                            & SeeSR / +StructSR    &  22.36 / 23.18 $\uparrow$   & 0.5673 / 0.5986 $\uparrow$    &  0.3194 / 0.3066 $\downarrow$    &  68.67 / \textbf{69.26} $\uparrow$    &  0.6934 / \textbf{0.6992} $\uparrow$       \\ \hline
\multirow{8}{*}{\makecell{RealSR \\ (real-world)}}     & BSRGAN           &  25.06    & 0.7401     &  \textbf{0.2656}     &  63.28     &  0.5115        \\
                            & Real-ESRGAN      &  24.32    & 0.7352     &  0.2726     &  60.45     &  0.4521        \\
                            & FeMaSR           &  23.74    & 0.7087     &  0.2937     &  59.06     &  0.5408        \\
                            & LDL           &  23.76    & 0.7198     &  0.2750    &  59.28     &  0.4430        \\ \cline{2-7} 
                            & StableSR / +StructSR &  23.41 / 24.31 $\uparrow$   & 0.6824 / \textbf{0.7447} $\uparrow$    &  0.3001 / 0.2915 $\downarrow$    &  65.05 / 67.68 $\uparrow$    &  0.6210 / 0.6624 $\uparrow$       \\
                            & DiffBIR / +StructSR  &  23.67 / \textbf{25.09} $\uparrow$   & 0.6245 / 0.6938 $\uparrow$    &  0.3626 / 0.3610 $\downarrow$    &  64.66 / 63.93 $\downarrow$    &  0.6545 / 0.6487 $\downarrow$       \\
                            & PASD / +StructSR     &  23.86 / 24.49 $\uparrow$   & 0.6946 / 0.7328 $\uparrow$    &  0.2960 / 0.2904 $\downarrow$    &  65.13 / 69.31 $\uparrow$    &  0.5760 / \textbf{0.6995} $\uparrow$       \\
                            & SeeSR / +StructSR    &  23.83 / 24.25 $\uparrow$   & 0.6947 / 0.7159 $\uparrow$    &  0.3007 / 0.2989 $\downarrow$    &  69.81 / \textbf{70.76} $\uparrow$    &  0.6703 / 0.6868 $\uparrow$       \\ \hline
\multirow{8}{*}{\makecell{DRealSR \\ (real-world)}}    & BSRGAN           &  27.38    & 0.7740     &  0.2858     &  57.16     &  0.5091        \\
                            & Real-ESRGAN      &  27.29    & 0.7768     &  0.2819     &  54.27     &  0.4515        \\
                            & FeMaSR           &  25.55    & 0.7246     &  0.3156     &  53.71     &  0.5639        \\
                            & LDL           &  25.97    & 0.7667     &  \textbf{0.2791}     &  53.95     &  0.4474        \\ \cline{2-7} 
                            & StableSR / +StructSR &  26.88 / 27.86 $\uparrow$   & 0.7252 / \textbf{0.7937} $\uparrow$    &  0.3182 / 0.2967 $\downarrow$    &  57.81 / 60.96 $\uparrow$    &  0.6029 / 0.6524 $\uparrow$       \\
                            & DiffBIR / +StructSR  &  25.40 / \textbf{27.98} $\uparrow$   & 0.6226 / 0.7566 $\uparrow$    &  0.4381 / 0.3640 $\downarrow$    &  60.37 / 59.76 $\downarrow$    &  0.6379 / 0.6176 $\downarrow$       \\
                            & PASD / +StructSR     &  26.48 / 27.14 $\uparrow$   & 0.7321 / 0.7662 $\uparrow$    &  0.3327 / 0.3151 $\downarrow$    &  58.90 / 65.33 $\uparrow$    &  0.5909 / \textbf{0.7243} $\uparrow$       \\
                            & SeeSR / +StructSR    &  26.74 / 27.17 $\uparrow$   & 0.7405 / 0.7754 $\uparrow$    &  0.3173 / 0.3028 $\downarrow$    &  65.09 / \textbf{66.80} $\uparrow$    &  0.6912 / 0.7013 $\uparrow$       \\ 
                            \bottomrule
\end{tabular}
\caption{Quantitative comparison with state-of-the-art methods on both synthetic and real-world benchmarks. BSRGAN (ICCV2021), Real-ESRGAN (ICCV2021), FeMaSR(ACM Multimedia2022) ,and LDL (CVPR2022) are GAN-based methods. StableSR (IJCV2024), DiffBIR (arXiv2023), PASD (ECCV2024), and SeeSR (CVPR2024) are diffusion-based methods. The best results of each metric are highlighted. Increasing and decreasing metrics are indicated by $\uparrow$ and $\downarrow$, respectively.}
\label{tab_cmp}
\end{table*}

\begin{figure*}[t!]
\centering
\includegraphics[width=0.98\textwidth]{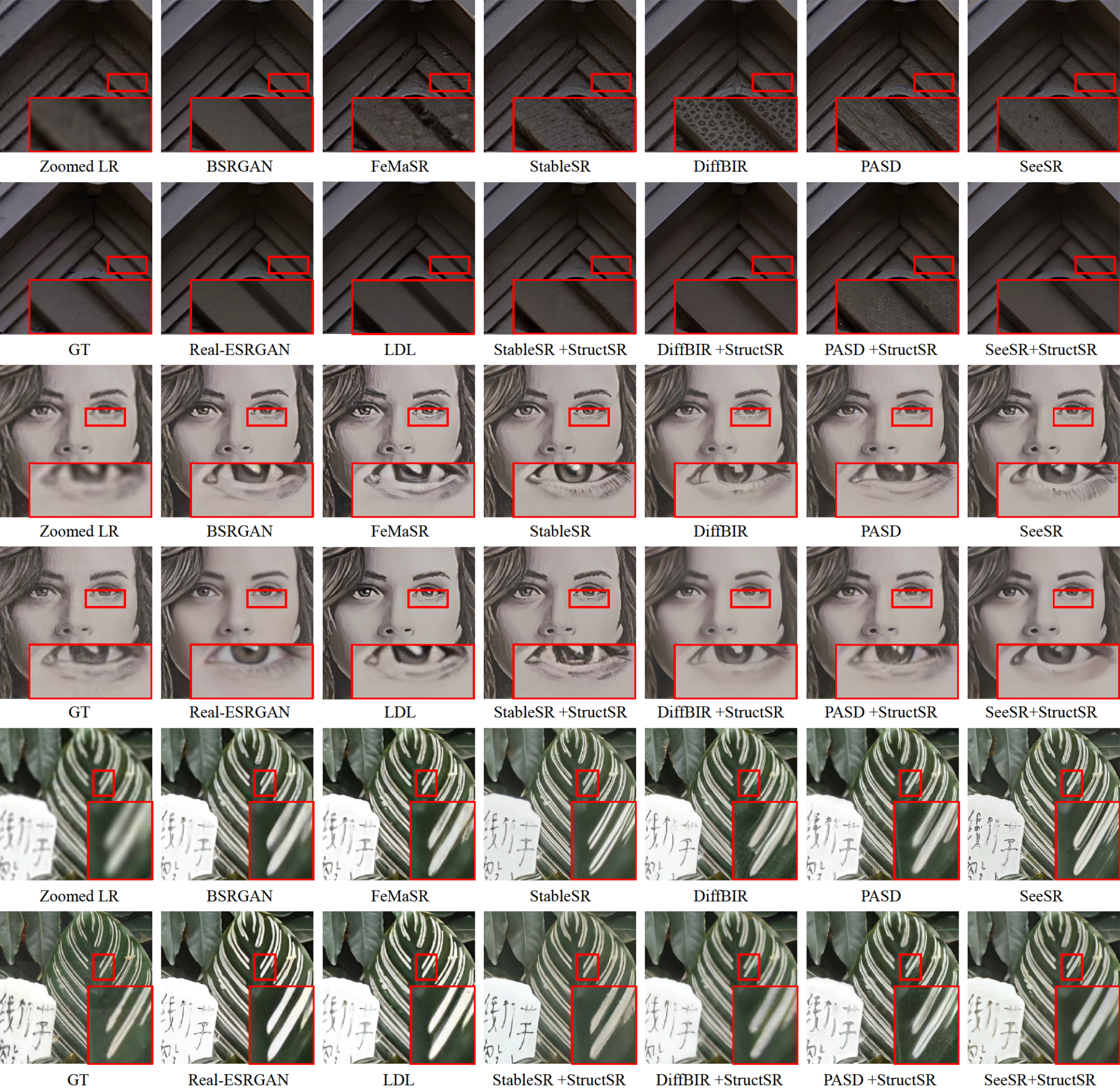}
\caption{Qualitative comparisons of different Real-ISR methods. Integration with StructSR significantly reduces spurious details of diffusion-based methods, resulting in high fidelity.
}
\label{fig_result}
\end{figure*}

\noindent \textbf{Qualitative Comparisons.} 
Fig. \ref{fig_result} shows the visual comparison of different Real-ISR methods.
From the first example, we can observe that BSRGAN, Real-ESRGAN, and LDL preserve the structures in LR images well. 
FeMaSR produces many artifacts due to the weakness of denoising ability. 
StableSR, PASD, and SeeSR produce a few spurious details. 
DiffBIR changes the structural information and generates spurious details.
Compared with the baselines, StructSR suppresses the generation of spurious details and corrects the structural errors.
From the second example, we can observe that the GAN-based methods lack the ability to generate details. Although the diffusion-based methods generate more details, they change the structural information and generate details inconsistent with GT. 
StructSR suppresses the spurious information compared to baselines.
The last example shows similar conclusions. 
These results demonstrate that StructSR can effectively enhance the structural fidelity and suppress the spurious details for diffusion-based Real-ISR. 
More comparisons are shown in the supplementary material.

\noindent \textbf{User Study.} To further validate the effectiveness of our approach, we conducted user studies on both synthetic and real-world data. 
Considering that the LR-HR image pairs in real-world data cannot be exactly matched like those in synthetic data, we use different settings for them.
On the synthetic data, followed the way in SR3~\cite{saharia2022image} and SeeSR~\cite{wu2024seesr}, participants were shown an LR image placed between two HR images: one was GT and the other was the Real-ISR output of a certain method. 
They were asked to decide, “Which HR image has the structure that better corresponds to the LR image?” 
When making their decision, participants were asked to consider the structural similarity to the LR image. 
A confusion rate was then calculated, which indicated whether participants preferred the GT or the Real-ISR output. 
On the real-world data, participants were shown an LR image along with all Real-ISR outputs and asked to answer “Which image has the structure that best corresponds to the LR image?” 
Then, the best rate was calculated to represent the probability of a model being selected.

We invite 30 participants to test representative Real-ISR methods. There are 12 synthetic test sets and 36 real-world test sets. Synthetic data are randomly sampled from {DIV2K-Val}, and real-world data are randomly sampled from {DRealSR and RealSR}. 
Each of the 30 participants is asked to make 144 (12x12) choices on synthetic data and 36 choices on real-world data. 
As shown in Table \ref{tab:user}, methods Integrated with our StructSR significantly outperform other methods in terms of selection rate on both synthetic and real-world data. 
In the user study on synthetic data, SeeSR achieves the best results among all methods without applying StructSR, while applying StructSR achieves a 58.3\% confusion rate, 25\% higher than the original. 
In the user study on real-world data, SeeSR applying our StructSR achieves the best selection rate of 16.7\%.

\begin{table}[t]
\setlength{\tabcolsep}{1mm}
\begin{tabular}{c|c|c}
\toprule
Methods   & Confusion rates &  Best rates \\ \hline
BSRGAN & 8.3\%   & 5.6\%   \\
Real-ESRGAN & 16.7\%   & 5.6\%   \\
FeMaSR & 8.3\%   & 2.8\%   \\
LDL & 8.3\%   & 2.8\%   \\ \cline{1-3}
StableSR / +StructSR & 25.0\% / 33.3\%  & 8.3\% / 13.9\%  \\
DiffBIR / +StructSR  & 16.7\% / 25.0\% & 2.8\% / 5.6\%  \\
PASD / +StructSR     & 25.0\% / 41.67\% & 11.1\% / 13.9\%   \\
SeeSR / +StructSR    & 33.3\% / 58.3\% & 11.1\% / 16.7\%   \\ 
\bottomrule
\end{tabular}
\caption{User study on synthetic and real-world data. Confusion rates are calculated on synthetic data and best rates are calculated on real-world data.}
\label{tab:user}
\end{table}

\section{Conclusion}
We propose StructSR, a novel plug-and-play method that enhances structural fidelity and suppresses spurious details for diffusion-based Real-ISR without any forms of model fine-tuning, external model priors, or high-level semantic knowledge. By exploring the structural similarity between reconstructed images and the original low-resolution image during the inference process, we found that the reconstructed images with high structural similarity values can be used to guide the inference process. Our work makes significant progress in effectively leveraging the internal characteristics of the diffusion-based Real-ISR models to synthesize high-resolution images with higher structural fidelity and fewer spurious details, as demonstrated through extensive experimental evaluations. More details can be found in the supplementary material.

\section{Acknowledgments}
This work is supported in part by the National Natural Science Foundation of China under grant 62272229 and U2441285, the Natural Science Foundation of Jiangsu Province under grant BK20222012, and Shenzhen Science and Technology Program JCYJ20230807142001004.

\bibliography{aaai25}

\newpage
\onecolumn
\appendix
\begin{center}
    \textbf{\LARGE{Supplementary Material}}
\end{center}

In this supplementary material, we first present the ablation studies on Structure Condition Embedding (SCE) and Image Details Embedding (IDE). Then we present further analysis of the definition of the early inference stage $T_{SAS}$. 

\setcounter{figure}{0}
\setcounter{table}{0}
\section{Ablation Study}

\subsection{Effectiveness of SCE and IDE}
\begin{figure}[H]
\centering
\includegraphics[width=\textwidth]{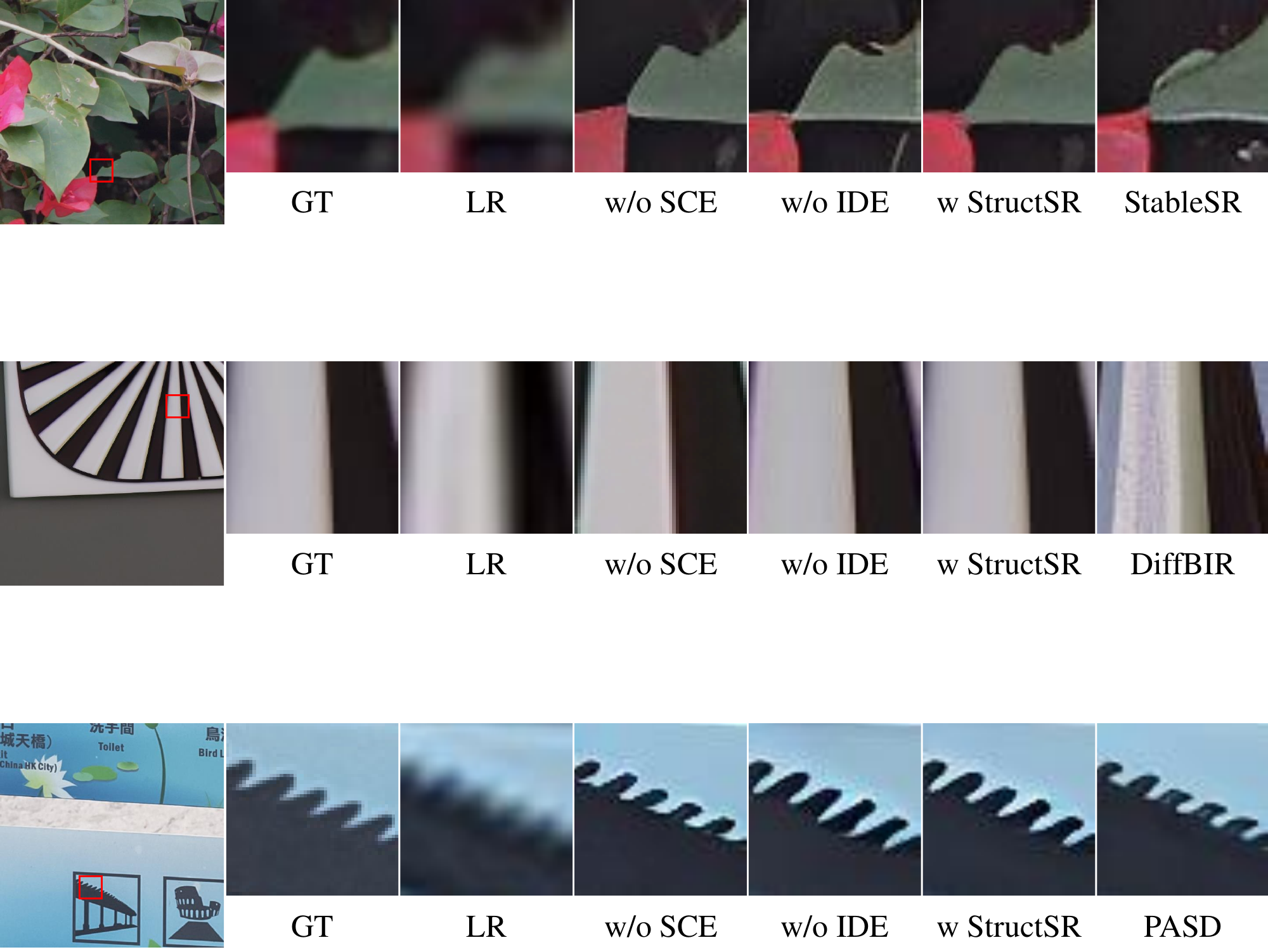}
\caption{Ablation study on SCE and IDE with state-of-the-art diffusion-based Real-ISR baselines. Integration with StructSR generates  high-fidelity structures by combining the clear structural guidance provided by SCE and the suppression of spurious details by IDE.}
\label{fig_ab}
\end{figure}

To demonstrate the effectiveness of the proposed Structure Condition Embedding (SCE) and Image Details Embedding (IDE) in improving the structural fidelity of the reconstructed image and suppressing spurious details, we perform several ablation studies. We used StableSR~\cite{wang2023exploiting}, DiffBIR~\cite{lin2023diffbir}, PASD~\cite{yang2023pixel}, and SeeSR~\cite{wu2024seesr} as baselines. The processed real-world datasets RealSR~\cite{realsr} and DRealSR~\cite{drealsr} were used for testing. The quantitative evaluation results of the ablation experiments are shown in Table~\ref{tab_ab}, and the corresponding visual performance is shown in Fig.~\ref{fig_ab}. 

Specifically, 1) \textbf{w/o StructSR}: Images are reconstructed only through the original inference process.
\textbf{2)} \textbf{w/o~SCE}: Images do not go through SCE, and only IDE is used for suppressing spurious details.
\textbf{3)} \textbf{w/o~IDE}: Images do not go through IDE, and only SCE is used for guiding noise prediction.
\textbf{4)} \textbf{w~StructSR}: Images are reconstructed by integrating StructSR into the original inference process.

As shown in Table~\ref{tab_ab} and Fig.~\ref{fig_ab}, for w/o~SCE, our proposed IDE effectively suppresses spurious details in the reconstructed image but causes the problem of over-smoothing. The results of w/o~IDE show that our proposed SCE provides clearer structural guidance for the inference process and helps the model produce more realistic details. In terms of metrics, compared with the baselines, w/o~SCE improves in PSNR and SSIM but decreases in LPIPS, MUSIQ, and CLIP-IQA. This shows that the results of w/o~SCE suppress spurious details while causing insufficient details in the reconstructed image because IDE suppresses the model's ability to generate details. Compared with the baselines, w/o~IDE improves in all metrics, but the improvement in PSNR is lower than those of w/o~SCE. This shows that the clear structural guidance provided by SCE guides the model in producing more details while guiding high-fidelity structural reconstruction, but some details are spurious.

\begin{table}[!h]
\setlength{\tabcolsep}{1mm}
\label{tab_ab}
\begin{tabular}{c|c|cccc|cccc}
\toprule
\hline
\multirow{2}{*}{Datasets}                     & \multirow{2}{*}{Metrics} & \multicolumn{4}{c|}{StableSR}                                                                                & \multicolumn{4}{c}{DiffBIR}                                                                                  \\ \cline{3-10} 
                                              &                          & \multicolumn{1}{c|}{w/o StructSR} & \multicolumn{1}{c|}{w/o SCE} & \multicolumn{1}{c|}{w/o IDE} & w StructSR & \multicolumn{1}{c|}{w/o StructSR} & \multicolumn{1}{c|}{w/o SCE} & \multicolumn{1}{c|}{w/o IDE} & w StructSR \\ \hline
\multicolumn{1}{c|}{\multirow{12}{*}{RealSR}} & PSNR $\uparrow$                     & \multicolumn{1}{c|}{23.41}        & \multicolumn{1}{c|}{24.29}   & \multicolumn{1}{c|}{24.07}   & \textbf{24.31}      & \multicolumn{1}{c|}{23.47}        & \multicolumn{1}{c|}{24.95}   & \multicolumn{1}{c|}{24.62}   & \textbf{25.09}      \\
\multicolumn{1}{c|}{}                         & SSIM $\uparrow$                     & \multicolumn{1}{c|}{0.6824}       & \multicolumn{1}{c|}{0.7001}  & \multicolumn{1}{c|}{0.7393}  & \textbf{0.7447}     & \multicolumn{1}{c|}{0.6245}       & \multicolumn{1}{c|}{0.6502}  & \multicolumn{1}{c|}{0.6846}  & \textbf{0.6938}     \\
\multicolumn{1}{c|}{}                         & LPIPS $\downarrow$                    & \multicolumn{1}{c|}{0.3001}       & \multicolumn{1}{c|}{0.3213}  & \multicolumn{1}{c|}{\textbf{0.2907}}  & 0.2915     & \multicolumn{1}{c|}{0.3626}       & \multicolumn{1}{c|}{0.3942}  & \multicolumn{1}{c|}{\textbf{0.3342}}  & 0.3610      \\
\multicolumn{1}{c|}{}                         & MUSIQ $\uparrow$                    & \multicolumn{1}{c|}{65.05}        & \multicolumn{1}{c|}{61.94}   & \multicolumn{1}{c|}{\textbf{70.29}}   & 67.68      & \multicolumn{1}{c|}{64.66}        & \multicolumn{1}{c|}{56.05}   & \multicolumn{1}{c|}{\textbf{65.74}}   & 63.93      \\
\multicolumn{1}{c|}{}                         & CLIP-IQA $\uparrow$                 & \multicolumn{1}{c|}{0.621}        & \multicolumn{1}{c|}{0.5698}  & \multicolumn{1}{c|}{\textbf{0.6967}}  & 0.6624     & \multicolumn{1}{c|}{0.6545}       & \multicolumn{1}{c|}{0.5336}  & \multicolumn{1}{c|}{\textbf{0.6815}}  & 0.6487     \\ \cline{2-10} 
\multicolumn{1}{c|}{}                         & \multirow{2}{*}{Metrics}        & \multicolumn{4}{c|}{PASD}                                                                                    & \multicolumn{4}{c}{SeeSR}                                                                                    \\ \cline{3-10} 
\multicolumn{1}{c|}{}                         &                          & \multicolumn{1}{c|}{w/o StructSR} & \multicolumn{1}{c|}{w/o SCE} & \multicolumn{1}{c|}{w/o IDE} & w StructSR & \multicolumn{1}{c|}{w/o StructSR} & \multicolumn{1}{c|}{w/o SCE} & \multicolumn{1}{c|}{w/o IDE} & w StructSR \\ \cline{2-10} 
\multicolumn{1}{c|}{}                         & PSNR $\uparrow$                     & \multicolumn{1}{c|}{23.86}        & \multicolumn{1}{c|}{24.26}   & \multicolumn{1}{c|}{23.91}   & \textbf{24.49}      & \multicolumn{1}{c|}{23.83}        & \multicolumn{1}{c|}{24.17}   & \multicolumn{1}{c|}{23.92}   & \textbf{24.25}      \\
\multicolumn{1}{c|}{}                         & SSIM $\uparrow$                     & \multicolumn{1}{c|}{0.6946}       & \multicolumn{1}{c|}{0.7022}  & \multicolumn{1}{c|}{0.7223}  & \textbf{0.7328}     & \multicolumn{1}{c|}{0.6947}       & \multicolumn{1}{c|}{0.7060}   & \multicolumn{1}{c|}{0.7114}  & \textbf{0.7159}     \\
\multicolumn{1}{c|}{}                         & LPIPS $\downarrow$                    & \multicolumn{1}{c|}{0.296}        & \multicolumn{1}{c|}{0.3071}  & \multicolumn{1}{c|}{\textbf{0.2901}}  & 0.2904     & \multicolumn{1}{c|}{0.3007}       & \multicolumn{1}{c|}{0.3065}  & \multicolumn{1}{c|}{\textbf{0.2971}}  & 0.2989     \\
\multicolumn{1}{c|}{}                         & MUSIQ $\uparrow$                    & \multicolumn{1}{c|}{65.13}        & \multicolumn{1}{c|}{68.55}   & \multicolumn{1}{c|}{\textbf{70.71}}   & 69.31      & \multicolumn{1}{c|}{69.81}        & \multicolumn{1}{c|}{65.23}   & \multicolumn{1}{c|}{\textbf{71.59}}   & 70.76      \\
\multicolumn{1}{c|}{}                         & CLIP-IQA $\uparrow$                 & \multicolumn{1}{c|}{0.5760}        & \multicolumn{1}{c|}{0.6730}   & \multicolumn{1}{c|}{\textbf{0.7077}}  & 0.6995     & \multicolumn{1}{c|}{0.6703}       & \multicolumn{1}{c|}{0.6409}  & \multicolumn{1}{c|}{\textbf{0.7054}}  & 0.6868     \\ \hline
\multicolumn{1}{c|}{}                         & \multirow{2}{*}{Metrics}        & \multicolumn{4}{c|}{StableSR}                                                                                    & \multicolumn{4}{c}{DiffBIR}                                                                                    \\ \cline{3-10} 
\multicolumn{1}{c|}{}                         &                          & \multicolumn{1}{c|}{w/o StructSR} & \multicolumn{1}{c|}{w/o SCE} & \multicolumn{1}{c|}{w/o IDE} & w StructSR & \multicolumn{1}{c|}{w/o StructSR} & \multicolumn{1}{c|}{w/o SCE} & \multicolumn{1}{c|}{w/o IDE} & w StructSR \\ \cline{2-10} 
\multicolumn{1}{c|}{\multirow{12}{*}{DRealSR}} & PSNR $\uparrow$                     & \multicolumn{1}{c|}{26.88}        & \multicolumn{1}{c|}{27.42}   & \multicolumn{1}{c|}{27.00}   & \textbf{27.86}      & \multicolumn{1}{c|}{25.40}        & \multicolumn{1}{c|}{27.17}   & \multicolumn{1}{c|}{26.61}   & \textbf{27.98}      \\
\multicolumn{1}{c|}{}                         & SSIM $\uparrow$                     & \multicolumn{1}{c|}{0.7252}       & \multicolumn{1}{c|}{0.7541}  & \multicolumn{1}{c|}{0.7827}  & \textbf{0.7937}     & \multicolumn{1}{c|}{0.6226}       & \multicolumn{1}{c|}{0.7396}  & \multicolumn{1}{c|}{0.7415}  & \textbf{0.7566}     \\
\multicolumn{1}{c|}{}                         & LPIPS $\downarrow$                    & \multicolumn{1}{c|}{0.3182}       & \multicolumn{1}{c|}{0.3259}  & \multicolumn{1}{c|}{\textbf{0.2958}}  & 0.2967     & \multicolumn{1}{c|}{0.4381}       & \multicolumn{1}{c|}{0.4649}  & \multicolumn{1}{c|}{\textbf{0.3556}}  & 0.364      \\
\multicolumn{1}{c|}{}                         & MUSIQ $\uparrow$                    & \multicolumn{1}{c|}{57.81}        & \multicolumn{1}{c|}{51.85}   & \multicolumn{1}{c|}{\textbf{64.25}}   & 60.96      & \multicolumn{1}{c|}{60.37}        & \multicolumn{1}{c|}{56.99}   & \multicolumn{1}{c|}{\textbf{61.06}}   & 59.76      \\
\multicolumn{1}{c|}{}                         & CLIP-IQA $\uparrow$                 & \multicolumn{1}{c|}{0.6029}       & \multicolumn{1}{c|}{0.5067}  & \multicolumn{1}{c|}{\textbf{0.7026}}  & 0.6524     & \multicolumn{1}{c|}{0.6379}       & \multicolumn{1}{c|}{0.4779}  & \multicolumn{1}{c|}{\textbf{0.6737}}  & 0.6176     \\ \cline{2-10} 
\multicolumn{1}{c|}{}                         & \multirow{2}{*}{Metrics}        & \multicolumn{4}{c|}{PASD}                                                                                    & \multicolumn{4}{c}{SeeSR}                                                                                    \\ \cline{3-10} 
\multicolumn{1}{c|}{}                         &                          & \multicolumn{1}{c|}{w/o StructSR} & \multicolumn{1}{c|}{w/o SCE} & \multicolumn{1}{c|}{w/o IDE} & w StructSR & \multicolumn{1}{c|}{w/o StructSR} & \multicolumn{1}{c|}{w/o SCE} & \multicolumn{1}{c|}{w/o IDE} & w StructSR \\ \cline{2-10} 
\multicolumn{1}{c|}{}                         & PSNR $\uparrow$                     & \multicolumn{1}{c|}{26.48}        & \multicolumn{1}{c|}{26.91}   & \multicolumn{1}{c|}{26.54}   & \textbf{27.14}      & \multicolumn{1}{c|}{26.74}        & \multicolumn{1}{c|}{26.95}   & \multicolumn{1}{c|}{26.81}   & \textbf{27.17}      \\
\multicolumn{1}{c|}{}                         & SSIM $\uparrow$                     & \multicolumn{1}{c|}{0.7321}       & \multicolumn{1}{c|}{0.7449}  & \multicolumn{1}{c|}{0.7575}  & \textbf{0.7662}     & \multicolumn{1}{c|}{0.7405}       & \multicolumn{1}{c|}{0.7661}  & \multicolumn{1}{c|}{0.7697}  & \textbf{0.7754}     \\
\multicolumn{1}{c|}{}                         & LPIPS $\downarrow$                    & \multicolumn{1}{c|}{0.3327}       & \multicolumn{1}{c|}{0.3418}  & \multicolumn{1}{c|}{\textbf{0.3122}}  & 0.3151     & \multicolumn{1}{c|}{0.3173}       & \multicolumn{1}{c|}{0.3232}  & \multicolumn{1}{c|}{\textbf{0.3006}}  & 0.3028     \\
\multicolumn{1}{c|}{}                         & MUSIQ $\uparrow$                    & \multicolumn{1}{c|}{58.9}         & \multicolumn{1}{c|}{65.12}   & \multicolumn{1}{c|}{\textbf{67.75}}   & 65.33      & \multicolumn{1}{c|}{65.09}        & \multicolumn{1}{c|}{60.44}   & \multicolumn{1}{c|}{\textbf{67.11}}   & 66.8       \\
\multicolumn{1}{c|}{}                         & CLIP-IQA $\uparrow$                 & \multicolumn{1}{c|}{0.5909}       & \multicolumn{1}{c|}{0.7171}  & \multicolumn{1}{c|}{\textbf{0.7320}}   & 0.7243     & \multicolumn{1}{c|}{0.6912}       & \multicolumn{1}{c|}{0.6523}  & \multicolumn{1}{c|}{\textbf{0.7332}}  & 0.7013     \\ \hline \hline \toprule
\end{tabular}
\caption{Ablation study on SCE and IDE with state-of-the-art diffusion-based Real-ISR baselines.}
\end{table}

\newpage
\section{Definition of early inference stage}
The definition of $T_{SAS}$ in the early inference stage is determined by the time when the maximum SSIM value between the reconstructed images and the LR image occurs. In the main text, we obtain three images with different degradation degrees based on a real image and study the changes in SSIM between the reconstructed images and them during the inference process. 

To obtain more generalized conclusions, we use more real-world images for research. Specifically, we perform center-crop on the HR images in the RealSR dataset to obtain \textbf{100} real-world images. Following the settings in the main text, we apply a combination of downsampling (D), Gaussian Kernel blur (D + B), and JPEG compression (J) to obtain three sets of LR images with different degradation degrees. 

\begin{figure}[H]
\centering
\includegraphics[width=0.7\textwidth]{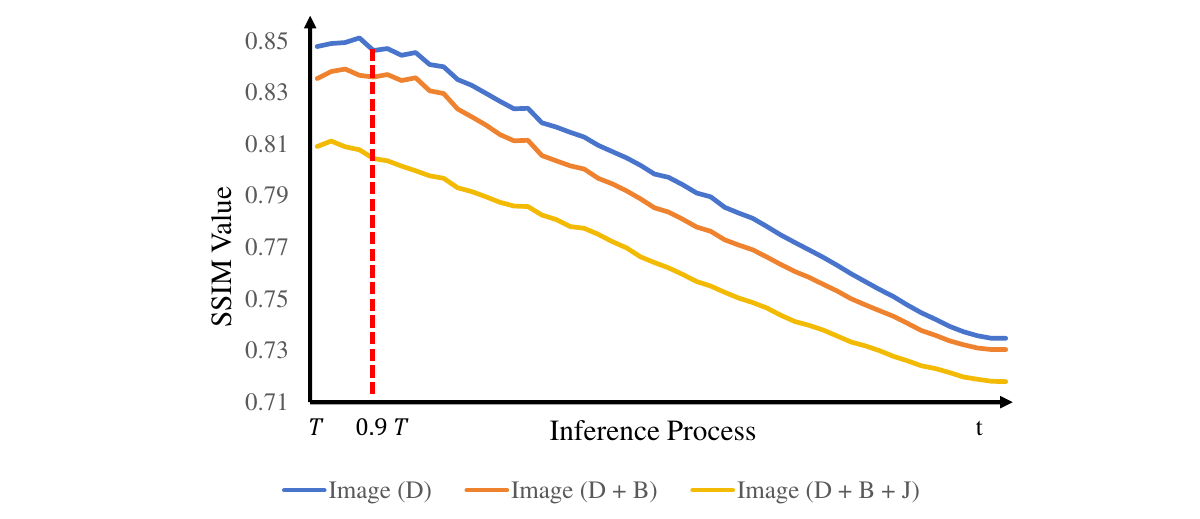}
\caption{Comparison of the SSIM between LR images with different degradation degrees and their temporal reconstructed images during the StableSR inference process with total inference timesteps $T = 200$.}
\label{fig_motivation3}
\end{figure}

As shown in Figure ~\ref{fig_motivation3}, by calculating the average SSIM value, we find that when $t\in[0.9\ T, T]$, the results of different degradation degrees all contain the maximum value. This shows that, in most cases, setting $T_{SAS}=0.1\ T$ can ensure that the reconstructed image screened by SAS is most consistent with the LR image in structure.

\begin{figure}[H]
\centering
\includegraphics[width=0.5\textwidth]{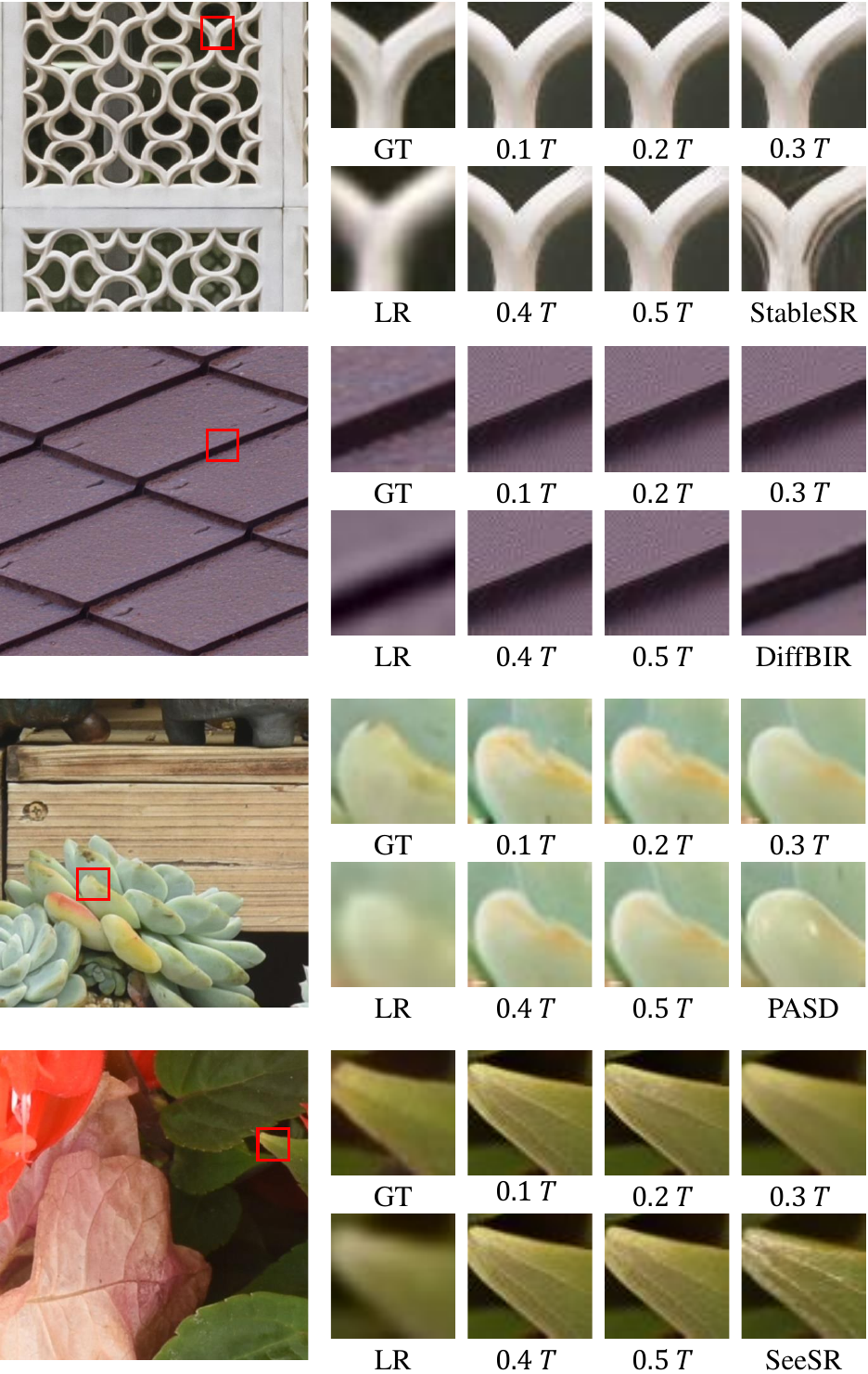}
\caption{Qualitative comparisons on different setting of $T_{SAS}$ with state-of-the-art diffusion-based Real-ISR baselines. Too-long early stage sacrifices the structural fidelity of reconstructed images. (Zoom in for details)}
\label{fig_ab_T}
\end{figure}

Based on the above findings, we set five sets of $T_{SAS}$ to verify the impact of longer early inference time steps on image quality. They are $T_{SAS}=0.1\ T$, $T_{SAS}=0.2\ T$, $T_{SAS}=0.3\ T$, $T_{SAS}=0.4\ T$, and $T_{SAS}=0.5\ T$. We use the processed RealSR~\cite{realsr} and DRealSR~\cite{drealsr} datasets as test datasets and StableSR~\cite{wang2023exploiting}, DiffBIR~\cite{lin2023diffbir}, PASD~\cite{yang2023pixel}, and SeeSR~\cite{wu2024seesr} as baselines. As shown in Table~\ref{tab_T} and Fig.~\ref{fig_ab_T}, for $T_{SAS}=0.3\ T$, integration with StructSR generates the most consistent structure with GT and the least spurious details. In terms of metrics, $T_{SAS}=0.3\ T$ achieves the best performance on PSNR, SSIM, and LPIPS. Compared with $T_{SAS}=0.3\ T$, the results of $T_{SAS}=0.4\ T$ and $T_{SAS}=0.5\ T$ contain more realistic but spurious details that do not match GT. In terms of metrics, $T_{SAS}=0.5\ T$ achieves the best performance on MUSIQ and CLIP-IQA, but is inferior to $T_{SAS}=0.3\ T$ in terms of PSNR, SSIM, and LPIPS. This indicates that a too-long early stage sacrifices the structural fidelity of the reconstructed images. By balancing structural fidelity and realistic details, we propose to adopt $T_{SAS}=0.3\ T$ as the early inference timesteps.

\begin{table}[!h]
\label{tab_T}
\begin{tabular}{c|c|ccccc|ccccc}
\toprule
\hline 
\multirow{2}{*}{Datasets}                     & \multirow{2}{*}{Metrics} & \multicolumn{5}{c|}{StableSR}                                                                                                       & \multicolumn{5}{c}{DiffBIR}                                                                                                                             \\ \cline{3-12} 
                                              &                          & \multicolumn{1}{c|}{$0.1\ T$} & \multicolumn{1}{c|}{$0.2\ T$} & \multicolumn{1}{c|}{$0.3\ T$} & \multicolumn{1}{c|}{$0.4\ T$} & $0.5\ T$ & \multicolumn{1}{c|}{$0.1\ T$} & \multicolumn{1}{c|}{$0.2\ T$} & \multicolumn{1}{c|}{$0.3\ T$} & \multicolumn{1}{c|}{$0.4\ T$} & \multicolumn{1}{c}{$0.5\ T$} \\ \hline
\multicolumn{1}{c|}{\multirow{12}{*}{RealSR}} & PSNR $\uparrow$                      & \multicolumn{1}{c|}{24.08}   & \multicolumn{1}{c|}{24.15}   & \multicolumn{1}{c|}{\textbf{24.31}}   & \multicolumn{1}{c|}{24.10}   & 24.06   & \multicolumn{1}{c|}{24.59}   & \multicolumn{1}{c|}{24.86}   & \multicolumn{1}{c|}{\textbf{25.09}}   & \multicolumn{1}{l|}{24.91}   & 24.83                       \\
\multicolumn{1}{c|}{}                         & SSIM $\uparrow$                     & \multicolumn{1}{c|}{0.7347}  & \multicolumn{1}{c|}{0.7411}  & \multicolumn{1}{c|}{\textbf{0.7447}}  & \multicolumn{1}{c|}{0.7387}  & 0.7215  & \multicolumn{1}{c|}{0.6847}  & \multicolumn{1}{c|}{0.6891}  & \multicolumn{1}{c|}{\textbf{0.6938}}  & \multicolumn{1}{l|}{0.6891}  & 0.6835                      \\
\multicolumn{1}{c|}{}                         & LPIPS $\downarrow$                    & \multicolumn{1}{c|}{0.3012}  & \multicolumn{1}{c|}{0.2974}  & \multicolumn{1}{c|}{\textbf{0.2915}}  & \multicolumn{1}{c|}{0.2948}  & 0.2967  & \multicolumn{1}{c|}{0.3683}  & \multicolumn{1}{c|}{0.3636}  & \multicolumn{1}{c|}{\textbf{0.3610}}   & \multicolumn{1}{l|}{0.3627}  & 0.3676                      \\
\multicolumn{1}{c|}{}                         & MUSIQ $\uparrow$                    & \multicolumn{1}{c|}{66.89}    & \multicolumn{1}{c|}{67.31}   & \multicolumn{1}{c|}{67.68}   & \multicolumn{1}{c|}{67.74}   & \textbf{67.86}   & \multicolumn{1}{c|}{63.28}   & \multicolumn{1}{c|}{63.69}   & \multicolumn{1}{c|}{63.93}   & \multicolumn{1}{l|}{64.05}   & \textbf{64.39}                       \\
\multicolumn{1}{c|}{}                         & CLIP-IQA $\uparrow$                 & \multicolumn{1}{c|}{0.6559}  & \multicolumn{1}{c|}{0.6605}  & \multicolumn{1}{c|}{0.6624}  & \multicolumn{1}{c|}{0.6632}  & \textbf{0.6643}  & \multicolumn{1}{c|}{0.6419}  & \multicolumn{1}{c|}{0.6468}  & \multicolumn{1}{c|}{0.6487}  & \multicolumn{1}{l|}{0.6489}  & \textbf{0.6494}                      \\ \cline{2-12} 
\multicolumn{1}{c|}{}                         & \multirow{2}{*}{Metrics} & \multicolumn{5}{c|}{PASD}                                                                                                           & \multicolumn{5}{c}{SeeSR}                                                                                                                               \\ \cline{3-12} 
\multicolumn{1}{c|}{}                         &                          & \multicolumn{1}{c|}{$0.1\ T$} & \multicolumn{1}{c|}{$0.2\ T$} & \multicolumn{1}{c|}{$0.3\ T$} & \multicolumn{1}{c|}{$0.4\ T$} & $0.5\ T$ & \multicolumn{1}{c|}{$0.1\ T$} & \multicolumn{1}{c|}{$0.2\ T$} & \multicolumn{1}{c|}{$0.3\ T$} & \multicolumn{1}{c|}{$0.4\ T$} & \multicolumn{1}{c}{$0.5\ T$} \\ \cline{2-12} 
\multicolumn{1}{c|}{}                         & PSNR $\uparrow$                     & \multicolumn{1}{c|}{23.97}   & \multicolumn{1}{c|}{24.13}   & \multicolumn{1}{c|}{\textbf{24.49}}   & \multicolumn{1}{c|}{24.15}   & 24.03   & \multicolumn{1}{c|}{23.90}   & \multicolumn{1}{c|}{24.04}   & \multicolumn{1}{c|}{\textbf{24.25}}   & \multicolumn{1}{l|}{24.07}   & 23.91                       \\
\multicolumn{1}{c|}{}                         & SSIM $\uparrow$                     & \multicolumn{1}{c|}{0.7259}  & \multicolumn{1}{c|}{0.7285}  & \multicolumn{1}{c|}{\textbf{0.7328}}  & \multicolumn{1}{c|}{0.7243}  & 0.7161  & \multicolumn{1}{c|}{0.7034}  & \multicolumn{1}{c|}{0.7124}  & \multicolumn{1}{c|}{\textbf{0.7159}}  & \multicolumn{1}{l|}{0.7114}  & 0.7079                      \\
\multicolumn{1}{c|}{}                         & LPIPS $\downarrow$                    & \multicolumn{1}{c|}{0.2966}  & \multicolumn{1}{c|}{0.2935}  & \multicolumn{1}{c|}{\textbf{0.2904}}  & \multicolumn{1}{c|}{0.2915}  & 0.2937  & \multicolumn{1}{c|}{0.3033}  & \multicolumn{1}{c|}{0.3023}  & \multicolumn{1}{c|}{\textbf{0.2989}}  & \multicolumn{1}{l|}{0.3004}  & 0.3027                      \\
\multicolumn{1}{c|}{}                         & MUSIQ $\uparrow$                    & \multicolumn{1}{c|}{67.98}   & \multicolumn{1}{c|}{68.94}   & \multicolumn{1}{c|}{69.31}   & \multicolumn{1}{c|}{69.38}   & \textbf{69.49}   & \multicolumn{1}{c|}{70.64}   & \multicolumn{1}{c|}{70.66}   & \multicolumn{1}{c|}{70.76}   & \multicolumn{1}{l|}{70.83}   & \textbf{70.90}                       \\
\multicolumn{1}{c|}{}                         & CLIP-IQA $\uparrow$                 & \multicolumn{1}{c|}{0.6948}  & \multicolumn{1}{c|}{0.6973}  & \multicolumn{1}{c|}{0.6995}  & \multicolumn{1}{c|}{0.6996}  & \textbf{0.7001}  & \multicolumn{1}{c|}{0.6837}  & \multicolumn{1}{c|}{0.6858}  & \multicolumn{1}{c|}{0.6868}  & \multicolumn{1}{l|}{0.6887}  & \textbf{0.6893}                      \\ \hline
\multicolumn{1}{c|}{}                         & \multirow{2}{*}{Metrics} & \multicolumn{5}{c|}{StableSR}                                                                                                           & \multicolumn{5}{c}{DiffBIR}                                                                                                                               \\ \cline{3-12} 
\multicolumn{1}{c|}{}                         &                          & \multicolumn{1}{c|}{$0.1\ T$} & \multicolumn{1}{c|}{$0.2\ T$} & \multicolumn{1}{c|}{$0.3\ T$} & \multicolumn{1}{c|}{$0.4\ T$} & $0.5\ T$ & \multicolumn{1}{c|}{$0.1\ T$} & \multicolumn{1}{c|}{$0.2\ T$} & \multicolumn{1}{c|}{$0.3\ T$} & \multicolumn{1}{c|}{$0.4\ T$} & \multicolumn{1}{c}{$0.5\ T$} \\ \cline{2-12} 
\multirow{12}{*}{DRealSR} & PSNR $\uparrow$                     & \multicolumn{1}{c|}{27.10}    & \multicolumn{1}{c|}{27.40}    & \multicolumn{1}{c|}{\textbf{27.86}}    & \multicolumn{1}{c|}{27.67}    & 26.46    & \multicolumn{1}{c|}{27.03}    & \multicolumn{1}{c|}{27.45}    & \multicolumn{1}{c|}{\textbf{27.98}}    & \multicolumn{1}{l|}{27.57}    & 27.39                        \\
                          & SSIM $\uparrow$                     & \multicolumn{1}{c|}{0.7833}   & \multicolumn{1}{c|}{0.7895}   & \multicolumn{1}{c|}{\textbf{0.7937}}   & \multicolumn{1}{c|}{0.7877}   & 0.7851   & \multicolumn{1}{c|}{0.7451}   & \multicolumn{1}{c|}{0.7494}   & \multicolumn{1}{c|}{\textbf{0.7566}}   & \multicolumn{1}{l|}{0.7475}   & 0.7428                       \\
                          & LPIPS $\downarrow$                    & \multicolumn{1}{c|}{0.3081}   & \multicolumn{1}{c|}{0.3002}   & \multicolumn{1}{c|}{\textbf{0.2967}}   & \multicolumn{1}{c|}{0.3037}   & 0.3043   & \multicolumn{1}{c|}{0.3787}   & \multicolumn{1}{c|}{0.3718}   & \multicolumn{1}{c|}{\textbf{0.3640}}    & \multicolumn{1}{l|}{0.3642}   & 0.3643                       \\
                          & MUSIQ $\uparrow$                    & \multicolumn{1}{c|}{60.18}    & \multicolumn{1}{c|}{60.57}    & \multicolumn{1}{c|}{60.96}    & \multicolumn{1}{c|}{61.14}    & \textbf{61.37}    & \multicolumn{1}{c|}{59.54}    & \multicolumn{1}{c|}{59.61}    & \multicolumn{1}{c|}{59.76}    & \multicolumn{1}{l|}{59.83}     & \textbf{59.91}                        \\
                          & CLIP-IQA $\uparrow$                 & \multicolumn{1}{c|}{0.6275}   & \multicolumn{1}{c|}{0.6396}   & \multicolumn{1}{c|}{0.6524}   & \multicolumn{1}{c|}{0.6558}   & \textbf{0.6582}   & \multicolumn{1}{c|}{0.6132}   & \multicolumn{1}{c|}{0.6165}   & \multicolumn{1}{c|}{0.6176}   & \multicolumn{1}{l|}{0.6197}   & \textbf{0.6209}                       \\ \cline{2-12} 
                          & \multirow{2}{*}{Metrics} & \multicolumn{5}{c|}{PASD}                                                                                                                & \multicolumn{5}{c}{SeeSR}                                                                                                                                    \\ \cline{3-12} 
                          &                          & \multicolumn{1}{c|}{$0.1\ T$} & \multicolumn{1}{c|}{$0.2\ T$} & \multicolumn{1}{c|}{$0.3\ T$} & \multicolumn{1}{c|}{$0.4\ T$} & $0.5\ T$ & \multicolumn{1}{c|}{$0.1\ T$} & \multicolumn{1}{c|}{$0.2\ T$} & \multicolumn{1}{c|}{$0.3\ T$} & \multicolumn{1}{c|}{$0.4\ T$} & \multicolumn{1}{c}{$0.5\ T$} \\ \cline{2-12} 
                          & PSNR $\uparrow$                     & \multicolumn{1}{c|}{26.70}    & \multicolumn{1}{c|}{26.93}    & \multicolumn{1}{c|}{\textbf{27.14}}    & \multicolumn{1}{c|}{27.05}    & 26.89    & \multicolumn{1}{c|}{26.83}    & \multicolumn{1}{c|}{27.09}    & \multicolumn{1}{c|}{\textbf{27.17}}    & \multicolumn{1}{l|}{27.03}    & 26.95                        \\
                          & SSIM $\uparrow$                     & \multicolumn{1}{c|}{0.7568}   & \multicolumn{1}{c|}{0.7604}   & \multicolumn{1}{c|}{\textbf{0.7662}}   & \multicolumn{1}{c|}{0.7615}   & 0.7572   & \multicolumn{1}{c|}{0.7681}   & \multicolumn{1}{c|}{0.7716}   & \multicolumn{1}{c|}{\textbf{0.7754}}   & \multicolumn{1}{l|}{0.7726}   & 0.7709                       \\
                          & LPIPS $\downarrow$                    & \multicolumn{1}{c|}{0.3365}   & \multicolumn{1}{c|}{0.3266}   & \multicolumn{1}{c|}{\textbf{0.3151}}   & \multicolumn{1}{c|}{0.3213}   & 0.3221   & \multicolumn{1}{c|}{0.3182}   & \multicolumn{1}{c|}{0.3083}   & \multicolumn{1}{c|}{\textbf{0.3028}}   & \multicolumn{1}{l|}{0.3054}   & 0.3073                       \\
                          & MUSIQ $\uparrow$                    & \multicolumn{1}{c|}{64.28}    & \multicolumn{1}{c|}{64.69}    & \multicolumn{1}{c|}{65.33}    & \multicolumn{1}{c|}{65.52}    & \textbf{65.56}    & \multicolumn{1}{c|}{65.95}    & \multicolumn{1}{c|}{66.03}    & \multicolumn{1}{c|}{66.80}     & \multicolumn{1}{l|}{66.87}    & \textbf{66.94}                         \\
                          & CLIP-IQA $\uparrow$                 & \multicolumn{1}{c|}{0.7203}   & \multicolumn{1}{c|}{0.7213}   & \multicolumn{1}{c|}{0.7243}   & \multicolumn{1}{c|}{0.7251}   & \textbf{0.7256}   & \multicolumn{1}{c|}{0.6969}   & \multicolumn{1}{c|}{0.7004}   & \multicolumn{1}{c|}{0.7013}   & \multicolumn{1}{l|}{0.7057}   & \textbf{0.7093}                       \\ \hline \hline \toprule
\end{tabular}
\caption{Quantitative experiments on different setting of $T_{SAS}$ with state-of-the-art diffusion-based Real-ISR baselines.}
\end{table}

\end{document}